%% file: Main.tex
\documentclass[letterpaper, 10 pt, conference]{ieeeconf}  

\IEEEoverridecommandlockouts                             

\usepackage{cite}
\usepackage[pdftex]{graphicx}
\usepackage{amsmath, amssymb}
\usepackage{array}
\usepackage[caption=false,font=footnotesize]{subfig}
\usepackage{siunitx}
\usepackage{threeparttable}

\usepackage{hyperref}
\usepackage[utf8]{inputenc}
\usepackage{textcomp, gensymb}

\usepackage{xcolor}
\usepackage{soul}
\usepackage{float}
\usepackage{booktabs}

\begin{document}
\input{1_HeadingsAndAll.tex}
\input{2_Abstract.tex}
\input{3_Introduction.tex}
\input{4_WorkingPrinciple.tex}
\input{6_ValveDesign.tex}
\input{7_TestBenchDesign.tex}
\input{8_ControlAlgorithm}
\input{9_ExperimentAndDiscussion.tex}
\input{10_Conclusion.tex}

\bibliographystyle{IEEEtran}
\bibliography{Main}
\end{document}

%% file: 1_HeadingsAndAll.tex
\title{\LARGE \bf

A Bimodal Hydrostatic Actuator for Robotic Legs with Compliant Fast Motion and High Lifting Force
}

\author{Alex Lecavalier$^{1}$,
Jeff Denis$^{1}$, Jean-Sébastien Plante$^{1}$,
Alexandre Girard$^{1}$\thanks{This work was supported by the Fonds québécois de la recherche sur la nature et les technologies (FRQNT) and the Natural Sciences and Engineering Research Council of Canada (NSERC).}\thanks{$^{1}$All authors are with the Department of Mechanical Engineering, Université de Sherbrooke, Qc, Canada.}}


\maketitle
\thispagestyle{empty}
\pagestyle{empty}

%% file: 2_Abstract.tex
\begin{abstract}

Robotic legs have bimodal operations: swing phases when the leg needs to move quickly in the air (high-speed, low-force) and stance phases when the leg bears the weight of the system (low-speed, high-force). Sizing a traditional single-ratio actuation system for such extremum operations leads to oversized heavy electric motor and poor energy efficiency, which hinder the capability of legged systems that bear the mass of their actuators and energy source. This paper explores an actuation concept where a hydrostatic transmission is dynamically reconfigured using valves to suit the requirements of each phase of a robotic leg. An analysis of the mass-delay-flow trade-off for the switching valve is presented. Then, a custom actuation system is built and integrated on a robotic leg test bench to evaluate the concept. Experimental results show that 1) small motorized ball valves can make fast transitions between operating modes when designed for this task, 2) the proposed operating principle and control schemes allow for seamless transitions, even during an impact with the ground and 3) the actuator characteristics address the needs of a leg bimodal operation in terms of force, speed and compliance.
\end{abstract}


%% file: 3_Introduction.tex
\section{Introduction}

\begin{figure}[h]
\vspace{-5pt}
\centering
\includegraphics[width=0.8\columnwidth]{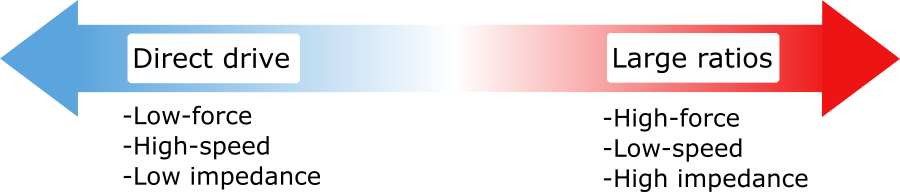}
\vspace{-10pt}
\caption{Trade-offs of geared motors with a fixed reduction ratio.}
\vspace{-5pt}
\label{fig_compromise}
\end{figure}

A robotic leg needs to quickly move through the air to reposition its foot, for instance, when stabilization or for fast gaits. Also, an ideal robotic leg have a small reflected inertia to limit the effects of the impact when the foot hits the ground. Lightly geared and direct-drive electric motors (EM) are thus well suited for those requirements and have been used for creating highly dynamic legged robots \cite{seok_actuator_2012} \cite{bledt_mit_2018}. On the other hand, in the stance phase, the leg must apply large forces to bear the weight of the robot and its payload. Without large reduction ratios, EM actuators exhibit poor torque density and efficiency at low-speed \cite{seok_actuator_2012} \cite{girard_two-speed_2015}. Thus, they are not well suited to stance phase requirements, especially if the robot needs to lift and carry heavy payloads. Alternatively, increasing the reduction ratios to meet the stance phase requirements will limit the maximum velocity and increase the inertia, thus penalizing the performance of the swing phase. These conflicting requirements for legs lead designers to compromise between multiple characteristics, illustrated in Figure~\ref{fig_compromise}, when using a fixed reduction ratio.

\begin{figure}[t]
\centering
\includegraphics[width=0.95\columnwidth]{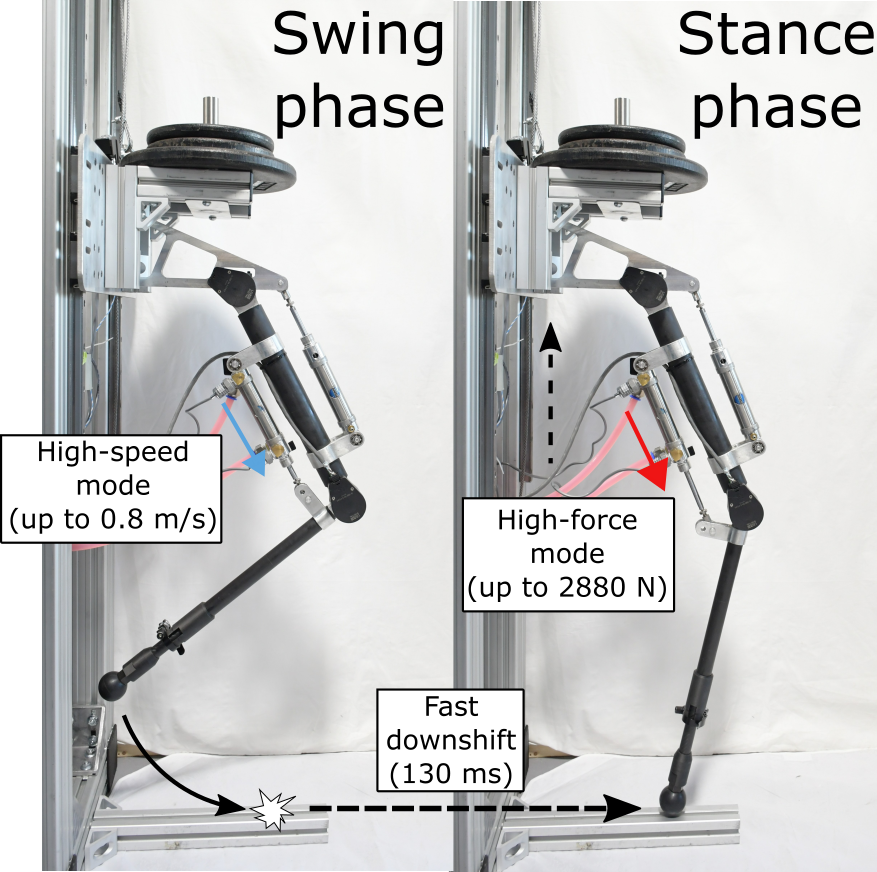}
\vspace{-10pt}
\caption{Bimodal demonstration on a robotic knee: swing phase (high-speed), stance phase (high-force).}
\vspace{-10pt}
\label{fig_fullLegExp}
\end{figure}%
    
Dynamically changing the reduction ratio, like most car powertrains, would allow a designer to avoid this performance compromise. If a robot leg actuator can downshift to a large reduction ratio during the stance phase, and upshift to a small reduction ratio for the swing phase, then the electric motor does not need to be oversized and would always work in an efficient operating range. 
Leveraging variable transmission actuators in this way have been sporadically explored by researchers in the field of robotics in the last decades. 
In this sense, Hirose relied on two parallel motors of different reduction ratio and an electromagnetic clutch to create a dual-mode transmission mechanism for an articulate prismatic leg \cite{hirose_design_1991}.
Bell proposed a dual-motor design for which the geared motor is electrically disconnected for high-speed motions to prevent back-emf power dissipation, but the geared motor inertia stay coupled to the output which limits the possible reduction ratios \cite{bell_two-motor_2020}.
Jeong~\textit{et al.} presented a single motor two-speed transmission based on twisted string actuation (TSA) and a dog clutch that is light and compact, but with many limitations in the operating conditions \cite{jeong_2-speed_2018}.
Lee~\textit{et al.} proposed a compact dual reduction actuator with a latching mechanism for a knee joint exoskeleton adapted for the walking phase and sit-to-stand phase, but without dynamic switching capabilities \cite{lee_flexible_2019}.
For seamless transitions, Jang~\textit{et al.} developed a continuously variable transmission (CVT) based on TSA, two motors and a differential gearbox, but is limited by the range of reduction ratios. 
Other serial dual-motor architectures were investigated too, requiring a differential and a brake that can be used to conduct seamless transitions, and shown the mass and energy advantages over single-ratio actuators \cite{girard_two-speed_2015} \cite{byeong-sang_kim_serial-type_2010} \cite{verstraten_kinematically_2019}. All in all, despite promising results, many challenges remain such as the trade-off between the complexity (and size) of the variable transmission and its ability to change the ratio in terms of range, speed, seamlessness and operating conditions in which it is possible to change the reduction ratio \cite{girard_fast_2017}.
A compact device that allows a fast and seamless switch between a small and a large reduction ratio in any operating conditions would be a breakthrough for many applications, especially for robotic legs.

This paper explores a novel two-speed hydrostatic architecture proposed in \cite{denis_multimodal_2022} which shown mass and energy advantages over a single-motor design. The concept leverages a hydrostatic transmission which is also beneficial to delocalize the motor of the moving linkages \cite{veronneau_lightweight_2019}\cite{veronneau_multifunctional_2020}\cite{khazoom_design_2019}\cite{khazoom_supernumerary_2020}. As illustrated in Fig.~\ref{fig_principle}, motorized ball valves are used to dynamically reconfigure the system between two operating modes tailored to the swing phase and stance phase. The concept is similar to the two-speed architecture explored by Verstraten \cite{verstraten_kinematically_2019} and Girard  \cite{girard_two-speed_2015}, but in the fluidic domain. The salient feature is that compact ball valves replace the high-force brake and the differential that were required in the mechanical domain, reducing the number of cumbersome components. Furthermore, the concept is more flexible in terms of conditions in which the system can downshift.
This paper presents an experimental assessment of the performance of this concept for actuating a robotic knee as shown in Figure~\ref{fig_fullLegExp}. The novel contributions are: 1) a mass-delay-flow tradeoff analysis for motorized ball valves used in the concept and 2) an experimental assessment of the ability of the concept to switch seamlessly underload, and 3) an experimental demonstration that the actuator prototype exhibit capabilities (force, speed and compliance) addressing the needs of robotic legs.
Section~\ref{section_workingPrinciple} presents the working principle and model of the proposed bimodal hydrostatic system. Section~\ref{section_valveDesign} discusses the use of hydraulic valves for reconfiguring the circuit and presents a motorized ball valve prototype. Section~\ref{section_testBenchDesign} presents the actuator prototype and the leg test bench. Section~\ref{section_control} presents control scheme used for coordinating the motor and valves, and experimental results with the prototype.

%% file: 4_WorkingPrinciple.tex
\section{Working principle and model}
\label{section_workingPrinciple}

The proposed two-speed architecture consists of a lightly geared electric motor (EM1) and a highly geared electric motor (EM2) which are respectively coupled to a high pitch and low pitch ball screw that actuates two hydraulic cylinders. Those two cylinders called master cylinder 1 (M1) and master cylinder 2 (M2) are connected to a slave cylinder on the leg, through a flexible hydraulic line. This results in a kinematically redundant system: the displacement of both master cylinders adds up to create a displacement at the slave cylinder (neglecting compressibility of the fluid) but the pressure is shared in the circuit. Additionally, two hydraulic valves can close the path to the lightly geared M1.

\begin{figure}[ht]
\centering
\vspace{-10pt}
\subfloat[High-speed mode;\centering]{\label{fig_HS}{\includegraphics{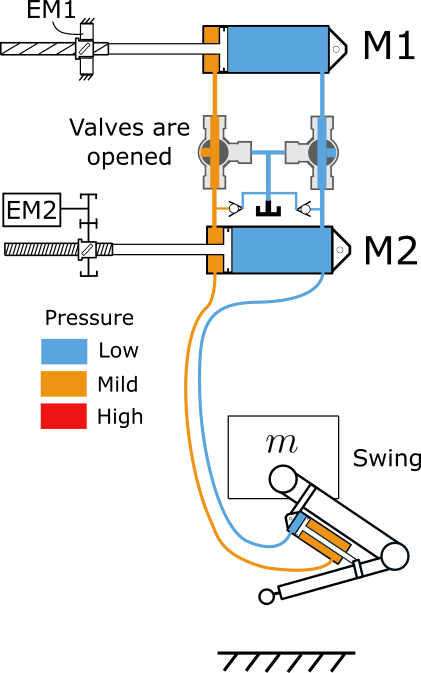}}}
\qquad
\subfloat[High-force mode;\centering]{\label{fig_HF}{\includegraphics{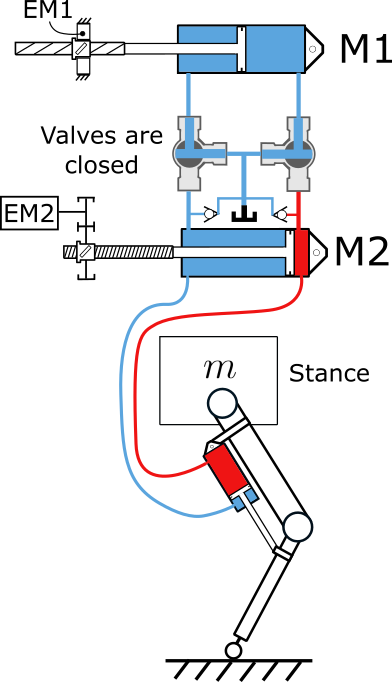}}}
\caption{Bimodal actuation principle of the proposed hydrostatic architecture.}
\vspace{-10pt}
\label{fig_principle}
\end{figure}%
    
\begin{figure}[ht]
\centering
\includegraphics[width=0.9\columnwidth]{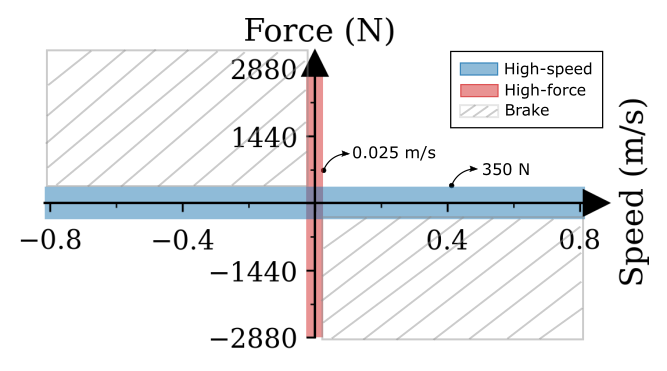}
\vspace{-10pt}
\caption{Bimodal system operating regions in terms of force and speed.}
\vspace{-5pt}
\label{fig_force_speed_quadrant}
\end{figure}

Hence, this architecture permits two main modes of operation: a high-speed mode (HS) for fast movement and backdrivability when the valves are opened and a high force mode (HF) when the valves are closed. When valves are open (Fig. \ref{fig_HS}), both M1 and M2 can contribute to the output motion (flow adds up, pressure is shared). This results in high-speed capability and a low reflected inertia at the output, but with force limited by EM1. When valves are closed (Fig. \ref{fig_HF}), only M2 contributes to the output (M1 is connected to the reservoir and can move freely). This results in high-force low-speed capabilities, as the configuration leads to a direct coupling of the highly geared piston to the slave piston. Furthermore, it is possible to generate large forces at high speeds in quadrants two and four (Fig. \ref{fig_force_speed_quadrant}) using partial opening of the valves to restrict the flow in order to brake the system. The operating modes capabilities in terms of force-speed are illustrated in the four quadrants at Figure \ref{fig_force_speed_quadrant}, using the specifications of the built actuator prototype. 

    
\begin{figure*}[ht]
\centering
\includegraphics[width=0.8\textwidth]{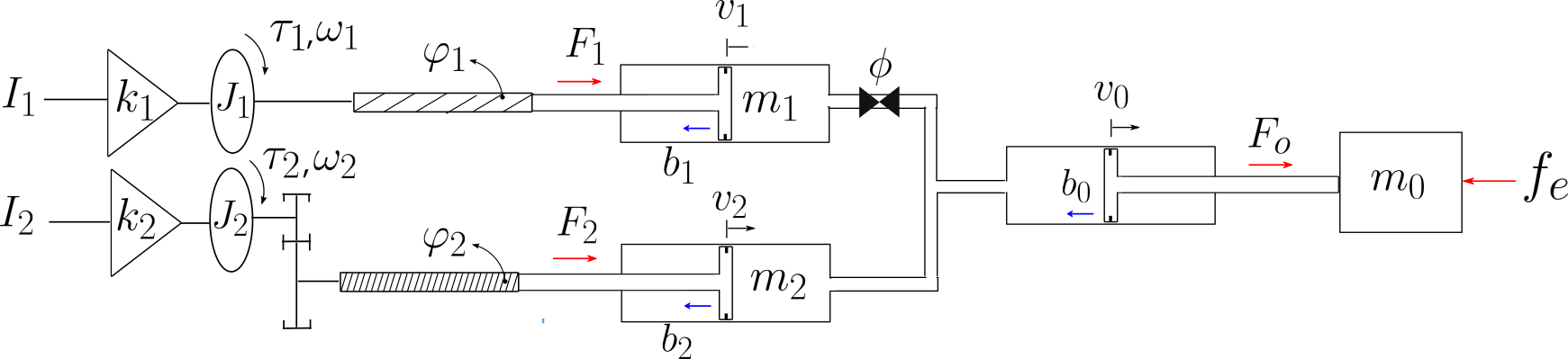}
\vspace{-10pt}
\caption{Lumped-parameter model of the proposed system.}
\vspace{-10pt}
\label{fig_lumped}
\end{figure*}
 
\subsection{Equations of motion}\label{section_EoM}

Here we present simplified equations of motions (EoM) describing the system behaviour, for all operating modes, based on lumped-parameter approach illustrated at Fig. \ref{fig_lumped}. Note that a single-action system is illustrated for simplicity. If we consider that the fluid in the circuit is incompressible, then the flow of M1 plus M2 equal to the incoming flow in the slave cylinder. When all cylinder area are equal, this leads to the following kinematic relationship between the pistons velocities and the motor velocities:
\begin{equation}
v_{o}=v_{1}+v_{2} = T_1^{-1} \omega_{1} + T_2^{-1}  \omega_{2} 
\label{eq_speed_relation}
\end{equation}
where $v_{o}$, $v_{1}$ and $v_{2}$ are the linear velocity of the output cylinder, M1 and M2,  $\omega_1$ and $\omega_2$ are the angular velocities of EM1 and EM2, and $T_1$ and $T_2$ are transformation ratio from motor angular motion to piston linear motion given by:
\begin{equation}
T_i = \frac{2\pi R_i}{\varphi_i}
\end{equation}
where $\varphi_i$ is the ball screw lead and $R_i$ is the reduction ratio between the motor shaft and the screw.

Considering the inertial properties associated with the moving parts, the passive dissipative forces and the active forces at all pistons, a 2 DoF dynamic model is constructed using as state variables the linear velocity of the output $\dot{v_o}$ and the linear velocity of M1 $\dot{v_1}$:
\begin{align}
\underbrace{ 
\mathbf{H}
\begin{bmatrix}
\dot{v_o}\\ \dot{v_1}
\end{bmatrix}
}_{Inertial}
+
\underbrace{ 
\begin{bmatrix}
b_2 + b_0 \\b_1 - b_2
\end{bmatrix}
}_{Disp.\;forces}
= 
\underbrace{
\mathbf{B}
\left[
\begin{array}{l}
 I_{1} \\
 I_{2} 
\end{array}
\right] 
}_{motors}
-
\underbrace{
\left[
\begin{array}{l}
 0 \\
 b( \phi )
\end{array}
\right] 
}_{throttling}
-
\underbrace{
\left[
\begin{array}{l}
 f_e \\
 0
\end{array}
\right] 
}_{ext.}
\label{eq_eom}
\end{align}
where
\begin{align}
\textbf{H} &= \left[
\begin{array}{c c}
 \scriptstyle m_o + m_2 +  J_2  T_2^2  & \scriptstyle - (m_2 + J_2 T_2^2) \\
 \scriptstyle - (m_2 + J_2 T_2^2).     & \scriptstyle m_1 +  J_1  T_1^2 + m_2 +  J_2  T_2^2
\end{array}
\right] 
\\
\mathbf{B} &= 
\left[
\begin{array}{c c}
\scriptstyle 0 & \scriptstyle k_2T_2 \\
\scriptstyle k_1T_1 & \scriptstyle -k_2 T_2
\end{array}
\right]
\end{align}
with $m_o$, $m_1$ and $m_2$, representing piston masses plus the inertial contribution of the transmission fluid in linear units at the piston; $J_1$ and $J_2$ representing the inertia of the motors; $b_o$, $b_1$ and $b_2$ representing speed-dependent dissipative forces acting respectively on the output, M1 and M2 pistons (friction in the seal, the ball screw and headloss in the fluid lines); $I_1$ and $I_2$ representing the electrical currents in the motors; $k_1$ and $k_2$ representing the motor torque constants; $f_e$ representing an external force acting on the output; and $b(\phi)$ representing a dissipative force acting on the M1 piston cause by closing partially the ball valve. This controllable dissipative force can be modelled with
\begin{equation}
b(\phi)=\frac{1}{2}sgn(v_1)k(\phi)v_1^2 \rho A 
\label{eq_force_local_loss}
\end{equation}
where $\phi$ is the valve angle, A is the bore area of the ball valve, $k(\phi)$ an experimental mapping between the ball angle and a loss coefficient, and $\rho$ is the fill fluid density.
    
During HF mode, M1 is disconnected and no flow from M1 contributes to the output motion. Setting $\dot{v}_1$ to zero in \eqref{eq_eom}, the model can be reduced to:
\begin{equation}
[m_o + \underbrace{m_2 + J_2 T_2^2}_{m_\text{A}\text{ for HF}}]
\dot{v}_o + b_o + b_2 = k_2 T_2 I_2 - f_e
\label{eq_hf}
\end{equation}
where $m_\text{A}$ is the reflected inertia due to the actuation.
For HS mode, both motor can contribute to the output motion. However, when isolating $\dot{v}_o$ in \eqref{eq_eom}, it can be seen that some terms related to the motion of M2 are negligible if $m_2 + J_2 T_2^2 \gg m_1 + J_1 T_1^2$
(which will be the case for the concept because $T_2$ is designed to be an order of magnitude greater than $T_1$), and the equation describing the motion of the output can be approximated by:
\begin{equation}
[m_o + \underbrace{m_1 + J_1 T_1^2}_{m_\text{A}\text{ for HS}}]
\dot{v}_o + b_o + b_1 = k_1 T_1 I_1 - b(\phi) - f_e
\label{eq_hs}
\end{equation}
All in all, with the approximation, the EoM of each mode have the same structure with the exception of an additional controllable dissipative force for HS mode. Equations show that the distinctive dynamic behaviour of each mode is fundamentally due to the mechanical advantage difference between $T_1$ and $T_2$.
%
Indeed, in HF mode with a large $T_2$, large forces can be applied (term $k_2 T_2 I_2$ of eq. \eqref{eq_hf}) and all terms not multiplied by $T_2$ in eq. \eqref{eq_hf} become negligible. The behaviour is mostly a motor fighting its own internal inertia (and friction) unafected by the load parameters and the external forces. 

%% file: 6_ValveDesign.tex
\section{Valve unit}
\label{section_valveDesign}
  
In order to meet the application requirements, the valve unit requires the following characteristics: 1) fast cycle, to allow quick switching to the HF mode after the leg contacts the ground, 2) small, in terms of mass and volume and 3) low-pressure drop when fully open, for efficiency and backdrivability of the HS mode. No off-the-shelf valves are designed for such requirements.

Servo ball valves (also butterfly) are commercially available, but they are designed for industrial applications and are very heavy, cumbersome and too slow to be used in a highly dynamic situation like a walking gait.
Fast servo valves have been used in mobile robots for their flow modulation accuracy, high bandwidth and lightness, for instance Boston Dynamics's robots \cite{noauthor_boston_2016}\cite{noauthor_atlas_nodate}. The main drawback of these servo valves is their inherent inefficiency and non-backdrivability since the modulation is based on restricting the flow \cite{buerger_novel_2010}.
Piloted cartridge valves have been used in digital hydraulics due to their small volume and their high speed of actuation. However, challenges remains on valve leakage and energy efficiency \cite{lantela_high-flow_2017}. Solenoid type cartridge valves have been used for a long time in automobile's anti-lock braking systems (ABS) \cite{linjama_digital_2011}. The downside is that solenoids tend to be inefficient, as they required a constant supply of energy to stay open, which generates thermal losses. While a permanent magnet latching solenoid can mitigate this issue, it is sensitive to vibration which is inherent to dynamic systems~\cite{solenoid_solutions_latching_2021}.

The concept in this paper leverages the hydraulics as a transmission only, i.e. a hydrostatic actuator, where the control modulation is done in the electric motor which is potentially much more efficient.
While industrial servo ball valves are not sufficient, the concept has demonstrated a potential use in robotic applications mainly because they are compact (with respect to the line diameter) and they exhibit a very low-pressure drop when fully open\cite{hashemi_experimentally_2020}. An analysis is thus conducted here to determine the viability of using servo ball valves for the application, and analyzing the trade-off between the size of the valve (mass), the pressure drop when open (bore diameter) and how fast can the valve open and close (cycle time). 

\begin{figure}[h]
\vspace{-10pt}
\centering
\includegraphics[width=0.8\columnwidth]{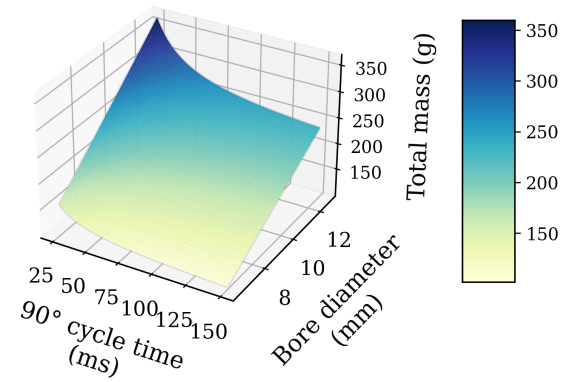}
\vspace{-10pt}
\caption{Mass mapping for aluminum motorized ball valves as a function of the bore diameter and 90° cycle time.}
\vspace{-10pt}
\label{fig_valveMap}
\end{figure}

Equation~\ref{eq_mass_total} estimates the mass of a servo ball valve as a function of the desired cycle time $\Delta t$ and its bore diameter~$d$, using a semi-empiric model given by:
\begin{equation}
\underbrace{
mass( d ,\Delta t ) 
}_{total}
=
\underbrace{
\frac{\pi\tau}{2\alpha\Delta t}
}_{motor}
+
\underbrace{\frac{\tau}{\beta}
}_{gearbox}
+
\underbrace{
\frac{\rho}{Re} \left[ \frac{Re}{\rho} \right]_{ref} m_{b}
}_{body}
\label{eq_mass_total}
\end{equation}
with
\begin{align}
\tau&\approx132d-0.2 
\label{eq_torque} \\
m_{b}&\approx41d-0.07
\label{eq_mass_brass}
\end{align}
This estimation is based on estimating individually the mass of the valve body, the electric motor actuating it and the gearbox connecting them. The mass of the electric motor is determined by the 90° cycle time $\Delta t$, the breakaway torque $\tau$ of the valve and electric motors specific power $\alpha$. Also, an empiric relationship between the breakaway torque and the bore diameter $d$ (eq. \ref{eq_torque}) based on experimental measurements on 6.35~mm, 9.52~mm and 12.7~mm commercial brass ball valves is used. The model assumes conservatively that the motor must apply the breakaway torque on the whole 90° stroke. 
The mass of the gearbox is estimated based on a torque-to-weight ratio. Specific power $\alpha$ is estimated to 600~W/kg and specific torque $\beta$ to 10~Nm/kg, both based on commercially available Maxon components in the range of 10 to 100~W. 
The mass of the valve body is estimated using a regression based on the valve bore diameter~$d$ (eq. \ref{eq_mass_brass}) using data on brass three-way ball valves (ranging from 6.35~mm to 19.05~mm) taken from the manufacturer's catalogue \cite{valworx_53685369_nodate}. A ratio of material specific strength is used to predict the mass of hypothetical optimal valves made of high-strength materials. Fig. \ref{fig_valveMap} shows the mass prediction when using aluminum 7075 instead of brass. Results show that custom servo ball valves could meet the requirements with reasonable mass.

\begin{figure}[h]
\centering
\vspace{-5pt}
\subfloat[Two three-way ball valves actuated with servo motors; \centering]{\label{fig_bv_prototype}{\includegraphics[width=0.4\columnwidth]{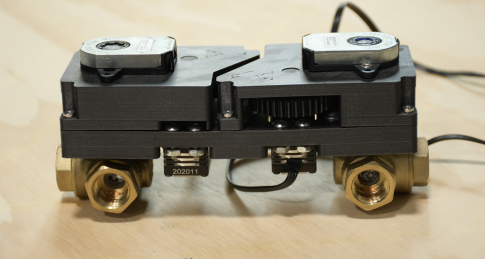}}}
\qquad
\subfloat[Mass distribution of the prototype;\centering]{\label{fig_pie}{\includegraphics[width=0.51\columnwidth]{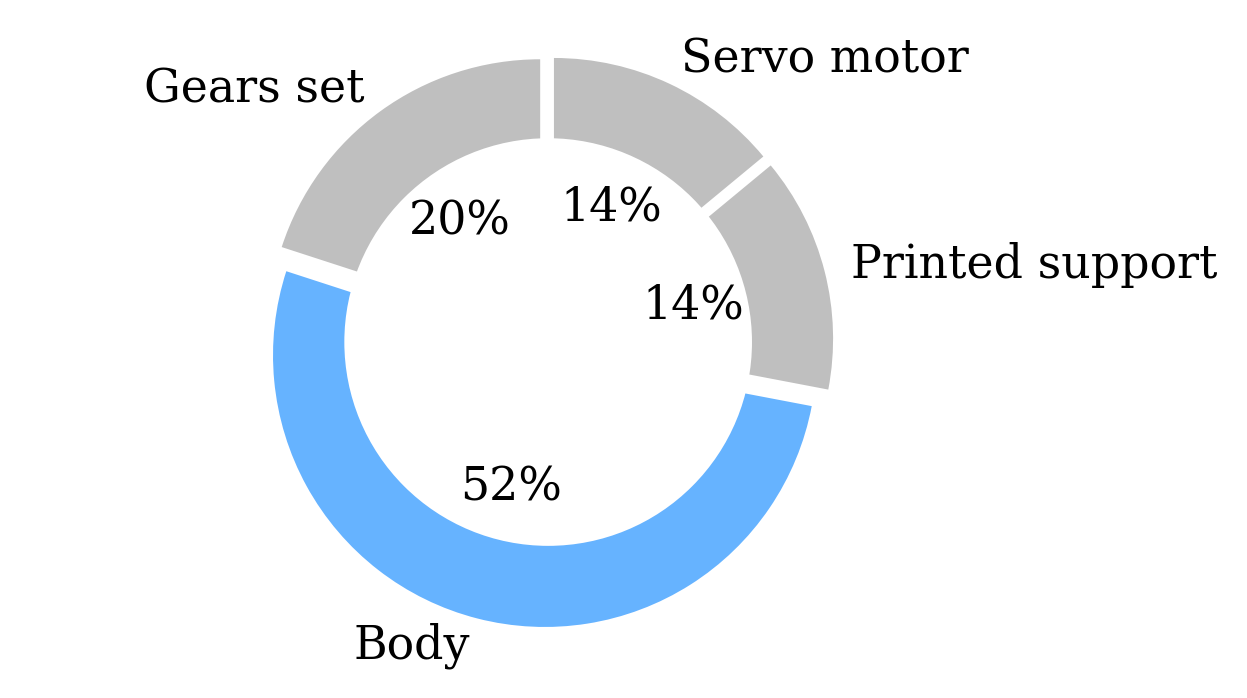}}}
\vspace{-5pt}
\caption{Motorized ball valves prototype built in our lab.}
\label{fig_built_prototype}
\end{figure}

Finally, to validate the concept, a prototype of two servo-actuated ball valves was designed and built, as shown in Figure \ref{fig_bv_prototype}. Standard brass T-pattern flow 3-way ball valves (Valworx 536903) were used and each valve is driven by a high torque servo motor (Savöx SB2262SG), through a custom single stage of spur gears for faster switching (1:1.4). The servo motor was chosen for its high torque (3.2~Nm), speed (16~rad.s$^{-1}$) and small volume (41 x 20 x 26~mm). 
The mass of the prototype is 415~g. As shown in Figure \ref{fig_pie}, the valve body accounts for 52\% of the total valve mass which means that choosing a light body material would decrease the mass significantly. 
The valve speed reaches a maximum value of 12.2~rad.s$^{-1}$ when opening and closing, hence a commutation time around 0.130~s. 
Using our empiric predictor equation eq. \eqref{eq_mass_total}, the estimated mass of an equivalent unit (9.52~mm bore, 0.130~s cycle time) but with a 7075 aluminum body is 172~g.


%% file: 7_TestBenchDesign.tex
\section{Actuator prototype and test bench}
\label{section_testBenchDesign}

\begin{figure}[h]
\vspace{-5pt}
\centering
\includegraphics[width=0.9\columnwidth]{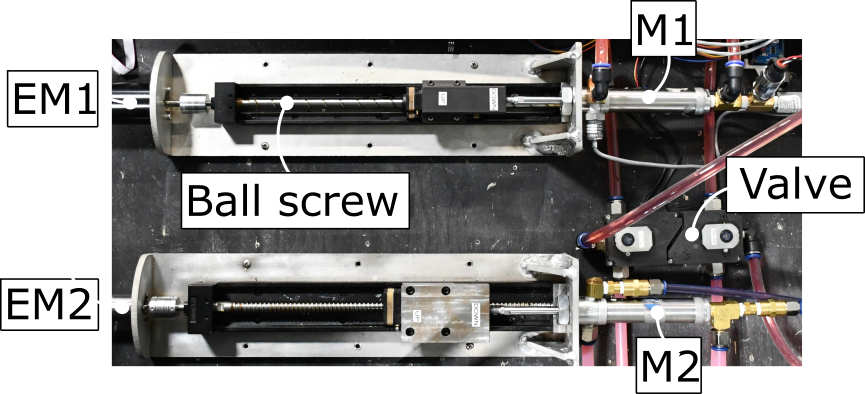}
\vspace{-10pt}
\caption{Hydrostatic test bench components layout.}
\vspace{-10pt}
\label{fig_test_bench_components}
\end{figure}

A prototype actuator using the custom valve unit and commercially available components was built and is shown in Figure \ref{fig_test_bench_components}. The prototype actuates a custom robotic leg test bench. One 200~W nominal motor (Maxon RE50, operated up to 2.6x the nominal current) and one 102~W nominal motor (Maxon DCX32L) are used for EM1 and EM2 respectively. EM1 is coupled to a 20~mm ball screw lead (NSK MCM05025H20K00) while EM2 is coupled to a 5~mm ball screw lead (NSK MCM06025H05K02) through a 28:1 planetary gear head (Maxon GPX32). Hydraulic cylinders (Bimba H-093-DUZ rated 3.45~MPa) are used and equipped with pressure sensors (MEAS MSP300). Propylene glycol is used as the transmission fluid for its non-toxicity, low viscosity and anti-corrosive properties. The test bench consists in a robotic leg attached to a cart (where steel plates can be installed as a payload) fixed to vertical linear guides (Fig.~\ref{fig_fullLegExp}). The bidirectional actuation of the knee can either swing the leg in the air, when the cart rests on the bottom of the linear guide, or lift the cart and payload  vertically when the foot touches the ground. The hip joint of the leg is unactivated. 
%

The prototype overall mechanical advantages are $T_1=315~\si{\per\meter}$ and $T_2=35186~\si{\per\meter}$, leading to the capabilities shown graphically on Fig. \ref{fig_force_speed_quadrant} and Table~\ref{table_performanceMap}. First, the computed reflected inertia reported at the output piston $m_A$ (see eq.~\ref{eq_hf} and \ref{eq_hs}) is two orders of magnitude lesser when using the HS mode, and less than the inertia of the lower leg structure ($m_o\approx17~\si{\kg}$ reported at the output cylinder when the leg is in the air). The maximum generated output force, using the maximum motor currents, is $350$~N in HS mode and $2880$~N in HF mode, which means the leg can sustain a payload of about 7.6~kg in HS mode and about 62~kg in HF mode. The maximum theoretical speed limit, based on $T_i^{-1} w_i$ with the maximum motors velocities are 0.8 $\si{\meter\per\second}$ in HS and 0.025 \si{\meter\per\second} in HF. Note that despite the high force and high-speed capabilities of the actuator, highly positive power motions such as jumping are not possible with the design. Finally, maximum acceleration capabilities are computed (using eq.~\ref{eq_hf} and \ref{eq_hs} and neglecting static friction) for two scenarios: 1) a swing phase starting from rest ($m_{o}\approx17~\si{\kg}$ and $f_e=0$) and 2) lifting a 25~kg payload starting from rest (inertia $m_{o}\approx460~\si{\kg}$ and $f_e$ is the payload gravity force reported to the slave piston). The mechanical advantage maximizing the acceleration would lead to $m_A=m_o$. Here we see that HS is well matched to accelerating the leg in the swing phase and that the reduction ratio of HF could be even larger if only considering maximization of the lifting phase acceleration.
\begin{table}[h!]
\vspace{-5pt}
\renewcommand{\arraystretch}{1.1}
\caption{Prototype actuator capabilities at the output piston}
\vspace{-5pt}
\label{table_performanceMap}
\centering
\begin{tabular}{ c c c c c c}
\hline 
 & $m_\text{A}$ & $F_\text{max}$ & $v_\text{max}$ & $\dot{v}_\text{max,swing}$ & $\dot{v}_\text{max,stance}$  \\
 Mode &  (\si{\kg}) & (\si{\newton}) & (\si{\meter\per\second}) & (\si{\meter\per\second\squared}) & (\si{\meter\per\second\squared}) \\
\hline 
\hline
HS & 10 & 350 & 0.8 & 11.5 & -1.1 \\
HF & 8900 & 2880 & 0.025 & 3.9 & 3.5 \\
\hline
\end{tabular}
\vspace{-5pt}
\end{table}

%% file: 8_ControlAlgorithm.tex
\section{Control and experiments}
\label{section_control}

\begin{figure}[h]
\centering
\includegraphics[width=\columnwidth]{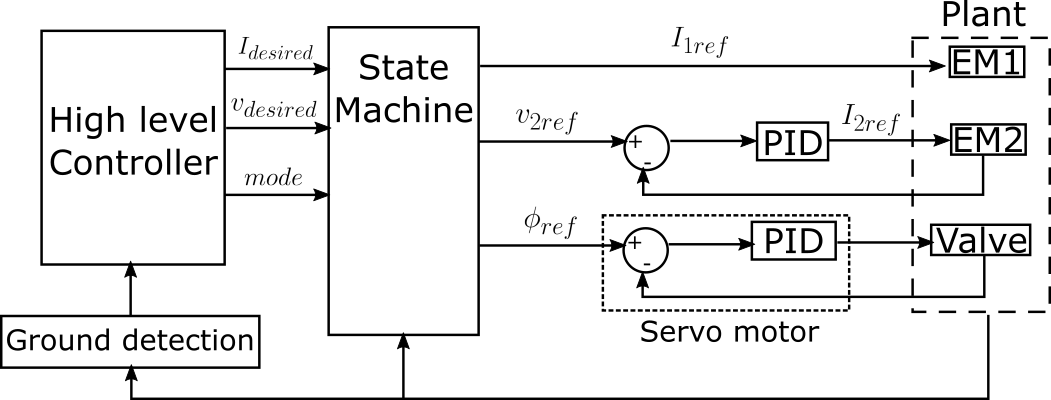}
\vspace{-10pt}
\caption{Control structure used for experiments.}
\vspace{-10pt}
\label{fig_control_loop}
\end{figure}

Robotic leg needs to continuously bear forces during the stance phase. Thus, when the leg hits the ground in HS mode and it is desired to switch to HF mode, the output force generated by the cylinder needs to be maintained during the transition, otherwise the leg may collapse. Fig. \ref{fig_control_loop} shows the control scheme proposed to handle the different operating modes and the challenges of smooth transitions.

\begin{figure}[h]
\centering
\includegraphics[width=\columnwidth]{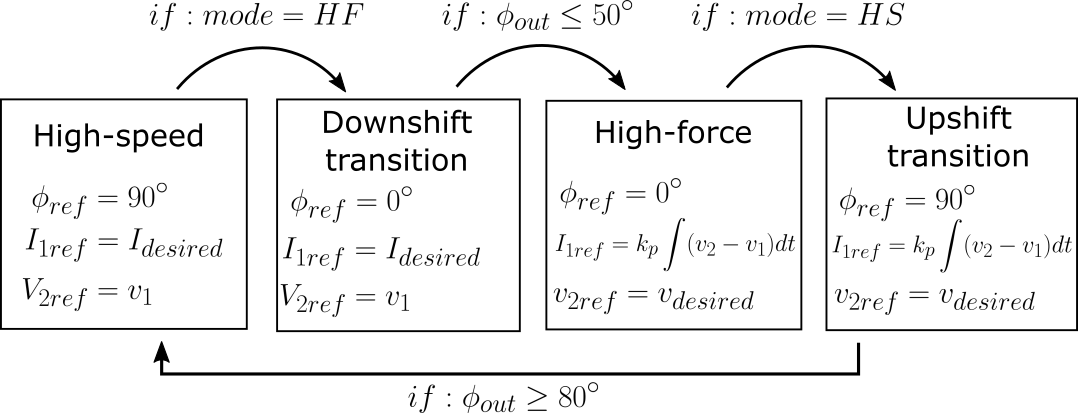}
\vspace{-10pt}
\caption{State machine.}
\vspace{-10pt}
\label{fig_state_machine}
\end{figure}

EM1 is controlled by imposing a current in the motor, which is closely related to the generated output pressure, since this line has a small mechanical advantage. EM2 is controlled by imposing its velocity with a low-level PID, since this motor mostly fights its own internal inertia and friction due to the large mechanical advantage. The valve is also controlled by imposing its position with the servo motor using its internal closed loop. Although there are two valves (for bidirectional actuation, see Fig. \ref{fig_principle}), only one variable is shown for simplicity, since both moves simultaneously. A high-level controller specifies the mode and a reference (current in HS and velocity in HF) to a state machine detailed in Fig. \ref{fig_state_machine}. For the purpose of demonstrating the capabilities of the actuator, the high-level controller is a hard-coded sequence of motion, with the exception of the downshift transition that is triggered by a ground contact detection using a combination of conditions on the knee joint encoder signal and the pressure signal in the slave cylinder.

Note that due to stroke limitations, it is desirable to have both master cylinders stroke as close as possible to the slave cylinder stroke. Thus, in HF mode, EM1 is commanded to follow M2 position and vice versa in HS mode.

%


%% file: 9_ExperimentAndDiscussion.tex

\subsection{Downshift and upshift between HS and HF modes}

\begin{figure}[h]
    \centering
    \includegraphics[width=\columnwidth]{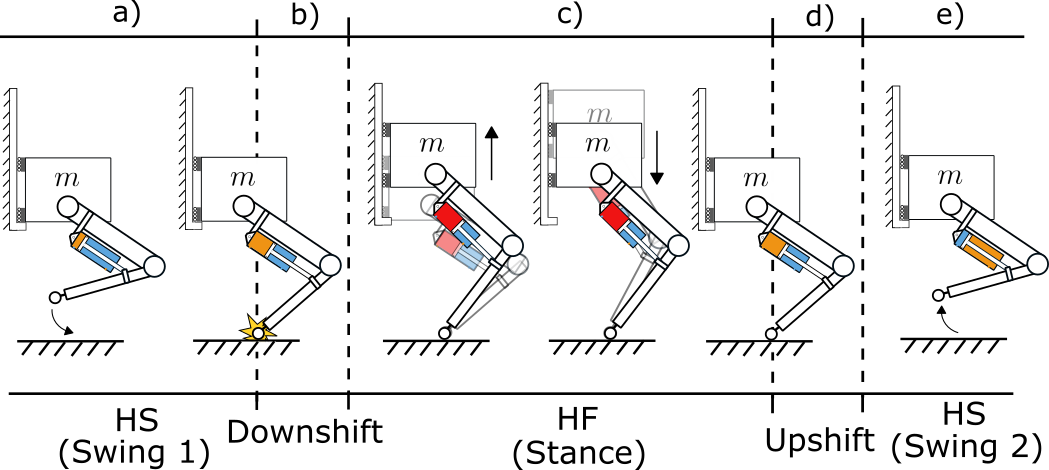}
    \vspace{-10pt}
    \caption{Motion sequence for the downshift and upshift transition test.}
    \vspace{-10pt}
    \label{fig_legExpSequence}
\end{figure}

\begin{figure}[h]
    \centering
    \includegraphics[width=0.8\columnwidth]{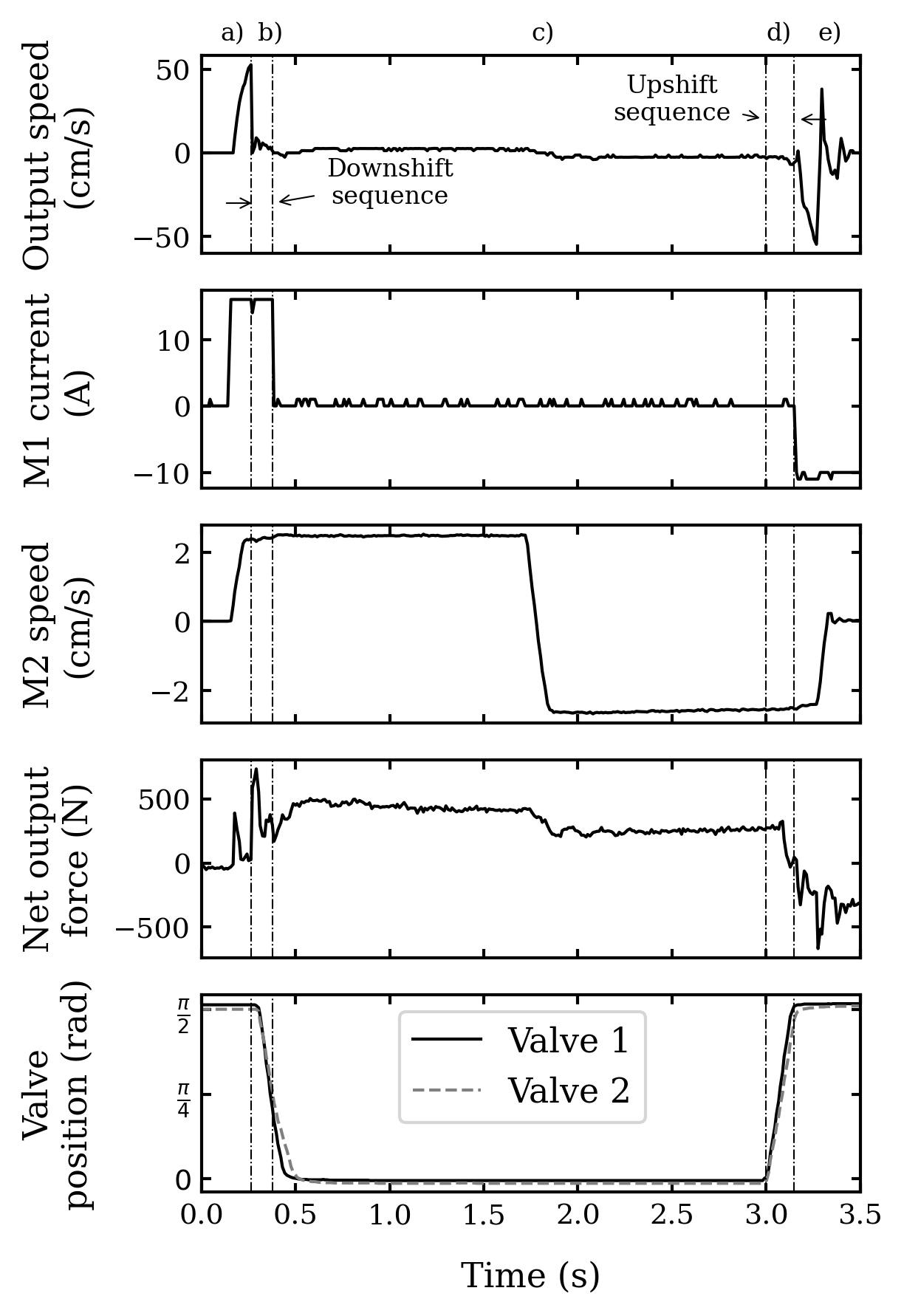}
    \vspace{-10pt}
    \caption{Results for the downshift and upshift transition test.}
    \vspace{-10pt}
    \label{fig_legExpTransitions}
\end{figure}

Fig. \ref{fig_legExpSequence} illustrates a complete test sequence analogous to a complete gait cycle. Starting from rest, the leg first swings at high-speed in HS mode (a) then the foot hits the ground at high-speed. The actuator detects this impact and downshifts in HF mode by closing the valves (b). Then, the leg moves up then down, lifting the $10~\si{\kg}$ payload (c). Finally, valves are opened to allow the leg to be retracted quickly using the HS mode. Control signals from the experiments are shown at Fig.~\ref{fig_legExpTransitions}. Note that the net output force signal (at output cylinder) is derived from the localized pressure sensor.
%
%
%
Results show that the downshift and upshift are conducted in about 0.130~s, which would allow mode dynamic mode transitions between swing and support phases for slow gait cycles. Also, it can be seen that during the downshift process, the output force does not drop and is maintained around 300~N, the capability of HS mode, until the valves are fully closed and EM2 can start bearing the whole weight. Pressure oscillations are observed, revealing that the compliance in the transmission (neglected in the model) leads to observable second order behaviours, but without affecting the system performance.  

\subsection{Throttling the valve for high braking forces}

\begin{figure}[h]
    \centering
    \includegraphics[width=0.8\columnwidth]{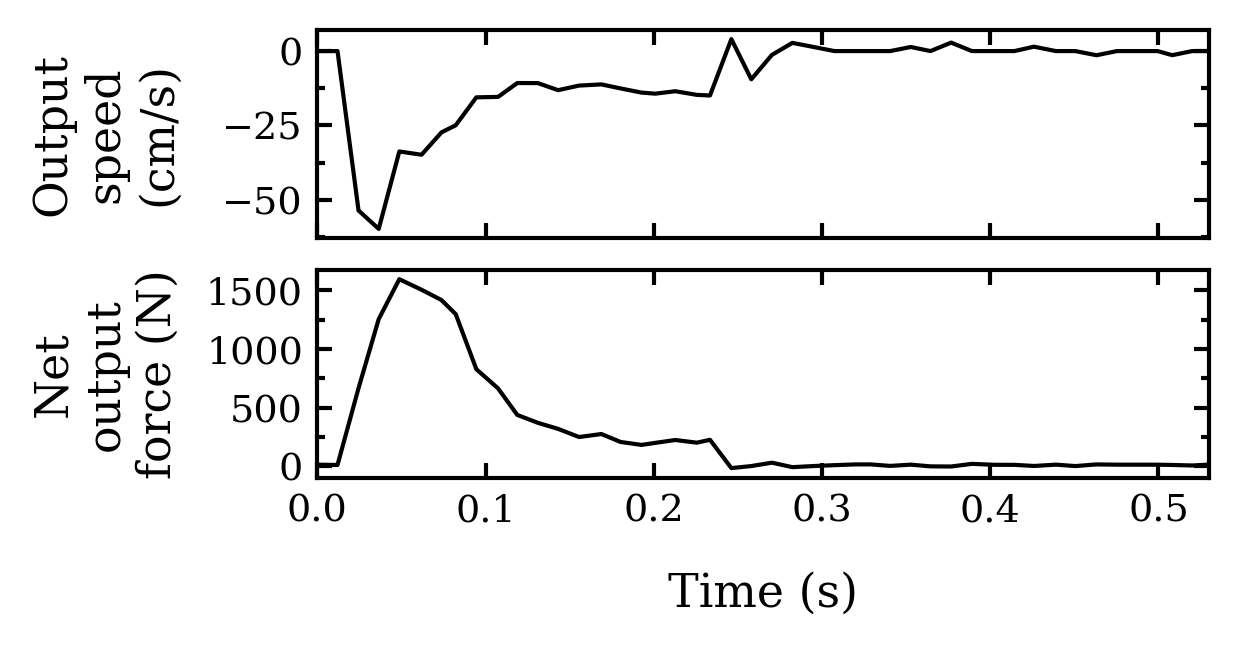}
    \vspace{-10pt}
    \caption{Drop test sequence showcasing the brake capacity of the valve.}
    \vspace{-10pt}
    \label{fig_damping}
\end{figure}

Previously proposed two-speed actuators had a limitation where if a large external force drove the system at high-speed, it would be impossible to downshift in this condition. Here it is demonstrated that the proposed actuator does not have this limitation since the valve itself can be used to create large braking forces (in quadrants II and IV see Fig. \ref{fig_force_speed_quadrant}) during HS mode. This feature would also be useful if a robotic system needed to dissipate high output kinetic energy for instance for a landing. 
Results for a 0.25~m drop test, see in Fig. \ref{fig_damping}, show that a braking force over 1500~N was generated at a high velocity during HS mode by throttling the ball valve, using the dissipative force $b(\phi)$ of eq. \eqref{eq_eom}. 
The robot is dropped with a total payload of 17~kg, including leg mass. Valves are halfway closed (constant 45° position) for dissipating energy. 
The peak braking power here is 960~W, which is largely over the maximal power of HS and HF modes. With proper characterization of the loss coefficient $k(\phi)$, it would be possible to dissipate even more energy by throttling the valve dynamically.


%% file: 10_Conclusion.tex
\section{Conclusion}




In this paper, a novel bimodal hydrostatic architecture that uses motorized ball valves to dynamically reconfigure the system between two operating modes of a robotic leg is presented. The equations of motion for the hydrostatic architecture are detailed. A custom switching valve unit is presented with an insight on the design trade-offs between the mass, delay and flow. Control strategies for upshift and downshift are detailed and tested on a robotic leg having completed a sequence similar to a walking gait. At last, the high-braking capabilities of the valves are demonstrated with a landing test.

Results show that 1) small motorized ball valves can make fast transitions between operating modes when designed for this task, 2) the proposed operating principle and control schemes allow for seamless transitions, even during an impact with the ground and 3) the actuator characteristics address the needs of a leg bimodal operation in terms of force, speed and compliance.
The general concept presented and tested could be applied to any system other than a robotic leg. However, many aspects of using bimodal hydrostatic architecture still need to be addressed. In terms of the proposed architecture, an optimized design of the proposed actuator, including the valve unit, is yet to be developed to prove that it is really suitable to embedded mobile robotic legs in terms of mass and for real gait cycles. Also, transition control could be fine-tuned based on pressure signals and force transmitted. Finally, an embedded version of the proposed hydrostatic architecture should be developed to replace the actual industrial components spread out on a table.




%% file: Main.bbl
\begin{thebibliography}{10}
\providecommand{\url}[1]{#1}
\csname url@samestyle\endcsname
\providecommand{\newblock}{\relax}
\providecommand{\bibinfo}[2]{#2}
\providecommand{\BIBentrySTDinterwordspacing}{\spaceskip=0pt\relax}
\providecommand{\BIBentryALTinterwordstretchfactor}{4}
\providecommand{\BIBentryALTinterwordspacing}{\spaceskip=\fontdimen2\font plus
\BIBentryALTinterwordstretchfactor\fontdimen3\font minus
  \fontdimen4\font\relax}
\providecommand{\BIBforeignlanguage}[2]{{%
\expandafter\ifx\csname l@#1\endcsname\relax
\typeout{** WARNING: IEEEtran.bst: No hyphenation pattern has been}%
\typeout{** loaded for the language `#1'. Using the pattern for}%
\typeout{** the default language instead.}%
\else
\language=\csname l@#1\endcsname
\fi
#2}}
\providecommand{\BIBdecl}{\relax}
\BIBdecl

\bibitem{seok_actuator_2012}
S.~Seok, A.~Wang, D.~Otten, and S.~Kim, ``Actuator design for high force
  proprioceptive control in fast legged locomotion,'' in \emph{2012
  {IEEE}/{RSJ} {International} {Conference} on {Intelligent} {Robots} and
  {Systems}}, Oct. 2012, pp. 1970--1975, iSSN: 2153-0866.

\bibitem{bledt_mit_2018}
G.~Bledt, M.~J. Powell, B.~Katz, J.~Di~Carlo, P.~M. Wensing, and S.~Kim,
  ``{MIT} {Cheetah} 3: {Design} and {Control} of a {Robust}, {Dynamic}
  {Quadruped} {Robot},'' in \emph{2018 {IEEE}/{RSJ} {International}
  {Conference} on {Intelligent} {Robots} and {Systems} ({IROS})}, Oct. 2018,
  pp. 2245--2252, iSSN: 2153-0866.

\bibitem{girard_two-speed_2015}
\BIBentryALTinterwordspacing
A.~Girard and H.~H. Asada, ``\BIBforeignlanguage{en}{A two-speed actuator for
  robotics with fast seamless gear shifting},'' in
  \emph{\BIBforeignlanguage{en}{2015 {IEEE}/{RSJ} {International} {Conference}
  on {Intelligent} {Robots} and {Systems} ({IROS})}}.\hskip 1em plus 0.5em
  minus 0.4em\relax Hamburg, Germany: IEEE, Sep. 2015, pp. 4704--4711.
  [Online]. Available: \url{http://ieeexplore.ieee.org/document/7354047/}
\BIBentrySTDinterwordspacing

\bibitem{hirose_design_1991}
\BIBentryALTinterwordspacing
S.~Hirose, K.~Yoneda, K.~Arai, and T.~Ibe, ``\BIBforeignlanguage{en}{Design of
  prismatic quadruped walking vehicle {TITAN} {VI}},'' in
  \emph{\BIBforeignlanguage{en}{Fifth {International} {Conference} on
  {Advanced} {Robotics} '{Robots} in {Unstructured} {Environments}}}.\hskip 1em
  plus 0.5em minus 0.4em\relax Pisa, Italy: IEEE, 1991, pp. 723--728 vol.1.
  [Online]. Available: \url{http://ieeexplore.ieee.org/document/240685/}
\BIBentrySTDinterwordspacing

\bibitem{bell_two-motor_2020}
\BIBentryALTinterwordspacing
J.~H. Bell, ``\BIBforeignlanguage{eng}{A two-motor actuator for legged robotics
  applications},'' Thesis, Massachusetts Institute of Technology, 2020,
  accepted: 2020-09-03T17:49:36Z. [Online]. Available:
  \url{https://dspace.mit.edu/handle/1721.1/127152}
\BIBentrySTDinterwordspacing

\bibitem{jeong_2-speed_2018}
\BIBentryALTinterwordspacing
S.~H. Jeong and K.-S. Kim, ``\BIBforeignlanguage{en}{A 2-{Speed} {Small}
  {Transmission} {Mechanism} {Based} on {Twisted} {String} {Actuation} and a
  {Dog} {Clutch}},'' \emph{\BIBforeignlanguage{en}{IEEE Robotics and Automation
  Letters}}, vol.~3, no.~3, pp. 1338--1345, Jul. 2018. [Online]. Available:
  \url{http://ieeexplore.ieee.org/document/8253814/}
\BIBentrySTDinterwordspacing

\bibitem{lee_flexible_2019}
\BIBentryALTinterwordspacing
Y.~Lee, J.~Lee, B.~Choi, M.~Lee, S.-g. Roh, K.~Kim, K.~Seo, Y.-J. Kim, and
  Y.~Shim, ``\BIBforeignlanguage{en}{Flexible {Gait} {Enhancing} {Mechatronics}
  {System} for {Lower} {Limb} {Assistance} ({GEMS} {L}-{Type})},''
  \emph{\BIBforeignlanguage{en}{IEEE/ASME Transactions on Mechatronics}},
  vol.~24, no.~4, pp. 1520--1531, Aug. 2019. [Online]. Available:
  \url{https://ieeexplore.ieee.org/document/8736810/}
\BIBentrySTDinterwordspacing

\bibitem{byeong-sang_kim_serial-type_2010}
\BIBentryALTinterwordspacing
{Byeong-Sang Kim}, {Jae-Bok Song}, and {Jung-Jun Park},
  ``\BIBforeignlanguage{en}{A {Serial}-{Type} {Dual} {Actuator} {Unit} {With}
  {Planetary} {Gear} {Train}: {Basic} {Design} and {Applications}},''
  \emph{\BIBforeignlanguage{en}{IEEE/ASME Transactions on Mechatronics}},
  vol.~15, no.~1, pp. 108--116, Feb. 2010. [Online]. Available:
  \url{http://ieeexplore.ieee.org/document/4840358/}
\BIBentrySTDinterwordspacing

\bibitem{verstraten_kinematically_2019}
\BIBentryALTinterwordspacing
T.~Verstraten, R.~Furnémont, P.~López-García, D.~Rodriguez-Cianca,
  B.~Vanderborght, and D.~Lefeber, ``\BIBforeignlanguage{en}{Kinematically
  redundant actuators, a solution for conflicting torque–speed
  requirements},'' \emph{\BIBforeignlanguage{en}{The International Journal of
  Robotics Research}}, vol.~38, no.~5, pp. 612--629, Apr. 2019. [Online].
  Available: \url{http://journals.sagepub.com/doi/10.1177/0278364919826382}
\BIBentrySTDinterwordspacing

\bibitem{girard_fast_2017}
\BIBentryALTinterwordspacing
A.~Girard, ``\BIBforeignlanguage{eng}{Fast and strong lightweight robots based
  on variable gear ratio actuators and control algorithms leveraging the
  natural dynamics},'' Thesis, Massachusetts Institute of Technology, 2017,
  accepted: 2017-10-04T14:47:02Z ISBN: 9781004236152. [Online]. Available:
  \url{https://dspace.mit.edu/handle/1721.1/111689}
\BIBentrySTDinterwordspacing

\bibitem{denis_multimodal_2022}
J.~Denis, A.~Lecavalier, J.-S. Plante, and A.~Girard, ``Multimodal
  {Hydrostatic} {Actuators} for {Wearable} {Robots}: {A} {Preliminary}
  {Assessment} of {Mass}-{Saving} and {Energy}-{Efficiency} {Opportunities},''
  in \emph{2022 {International} {Conference} on {Robotics} and {Automation}
  ({ICRA})}, May 2022, pp. 8112--8118.

\bibitem{veronneau_lightweight_2019}
\BIBentryALTinterwordspacing
C.~Veronneau, J.~Denis, L.-P. Lebel, M.~Denninger, J.-S. Plante, and A.~Girard,
  ``\BIBforeignlanguage{en}{A {Lightweight} {Force}-{Controllable} {Wearable}
  {Arm} {Based} on {Magnetorheological}-{Hydrostatic} {Actuators}},'' in
  \emph{\BIBforeignlanguage{en}{2019 {International} {Conference} on {Robotics}
  and {Automation} ({ICRA})}}.\hskip 1em plus 0.5em minus 0.4em\relax Montreal,
  QC, Canada: IEEE, May 2019, pp. 4018--4024. [Online]. Available:
  \url{https://ieeexplore.ieee.org/document/8793978/}
\BIBentrySTDinterwordspacing

\bibitem{veronneau_multifunctional_2020}
\BIBentryALTinterwordspacing
C.~Veronneau, J.~Denis, L.-P. Lebel, M.~Denninger, V.~Blanchard, A.~Girard, and
  J.-S. Plante, ``\BIBforeignlanguage{en}{Multifunctional {Remotely} {Actuated}
  3-{DOF} {Supernumerary} {Robotic} {Arm} {Based} on {Magnetorheological}
  {Clutches} and {Hydrostatic} {Transmission} {Lines}},''
  \emph{\BIBforeignlanguage{en}{IEEE Robotics and Automation Letters}}, vol.~5,
  no.~2, pp. 2546--2553, Apr. 2020. [Online]. Available:
  \url{https://ieeexplore.ieee.org/document/8962256/}
\BIBentrySTDinterwordspacing

\bibitem{khazoom_design_2019}
\BIBentryALTinterwordspacing
C.~Khazoom, C.~Veronneau, J.-P.~L. Bigue, J.~Grenier, A.~Girard, and J.-S.
  Plante, ``\BIBforeignlanguage{en}{Design and {Control} of a {Multifunctional}
  {Ankle} {Exoskeleton} {Powered} by {Magnetorheological} {Actuators} to
  {Assist} {Walking}, {Jumping}, and {Landing}},''
  \emph{\BIBforeignlanguage{en}{IEEE Robotics and Automation Letters}}, vol.~4,
  no.~3, pp. 3083--3090, Jul. 2019. [Online]. Available:
  \url{https://ieeexplore.ieee.org/document/8744625/}
\BIBentrySTDinterwordspacing

\bibitem{khazoom_supernumerary_2020}
\BIBentryALTinterwordspacing
C.~Khazoom, P.~Caillouette, A.~Girard, and J.-S. Plante,
  ``\BIBforeignlanguage{en}{A {Supernumerary} {Robotic} {Leg} {Powered} by
  {Magnetorheological} {Actuators} to {Assist} {Human} {Locomotion}},''
  \emph{\BIBforeignlanguage{en}{IEEE Robotics and Automation Letters}}, vol.~5,
  no.~4, pp. 5143--5150, Oct. 2020. [Online]. Available:
  \url{https://ieeexplore.ieee.org/document/9128040/}
\BIBentrySTDinterwordspacing

\bibitem{noauthor_boston_2016}
\BIBentryALTinterwordspacing
``\BIBforeignlanguage{en}{Boston {Dynamics}' {Marc} {Raibert} on {Next}-{Gen}
  {ATLAS}: "{A} {Huge} {Amount} of {Work}"},'' Feb. 2016, section: Robotics.
  [Online]. Available:
  \url{https://spectrum.ieee.org/boston-dynamics-marc-raibert-on-nextgen-atlas}
\BIBentrySTDinterwordspacing

\bibitem{noauthor_atlas_nodate}
\BIBentryALTinterwordspacing
``\BIBforeignlanguage{en}{Atlas™}.'' [Online]. Available:
  \url{https://www.bostondynamics.com/atlas}
\BIBentrySTDinterwordspacing

\bibitem{buerger_novel_2010}
\BIBentryALTinterwordspacing
S.~P. Buerger and N.~Hogan, ``\BIBforeignlanguage{en}{Novel {Actuation}
  {Methods} for {High} {Force} {Haptics}},''
  \emph{\BIBforeignlanguage{en}{Advances in Haptics}}, p.~33, 2010. [Online].
  Available:
  \url{http://www.intechopen.com/books/advances-in-haptics/novel-actuation-methods-for-high-force-haptics}
\BIBentrySTDinterwordspacing

\bibitem{lantela_high-flow_2017}
\BIBentryALTinterwordspacing
T.~Lantela and M.~Pietola, ``\BIBforeignlanguage{en}{High-flow rate miniature
  digital valve system},'' \emph{\BIBforeignlanguage{en}{International Journal
  of Fluid Power}}, vol.~18, no.~3, pp. 188--195, Sep. 2017. [Online].
  Available:
  \url{https://www.tandfonline.com/doi/full/10.1080/14399776.2017.1358025}
\BIBentrySTDinterwordspacing

\bibitem{linjama_digital_2011}
M.~Linjama, ``\BIBforeignlanguage{en}{Digital fluid power - {State} of the
  art},'' \emph{\BIBforeignlanguage{en}{The Twelfth Scandinavian International
  Conference on Fluid Power}}, vol.~12, no.~3, p.~23, May 2011.

\bibitem{solenoid_solutions_latching_2021}
\BIBentryALTinterwordspacing
S.~Solutions, ``Latching {Solenoid} {Valves} {Low} {Energy},'' 2021,
  https://www.solenoidsolutionsinc.com/specialty-valves/latching-solenoid-valves-low-energy/.
  [Online]. Available:
  \url{https://www.solenoidsolutionsinc.com/specialty-valves/latching-solenoid-valves-low-energy/}
\BIBentrySTDinterwordspacing

\bibitem{hashemi_experimentally_2020}
S.~Hashemi, H.~Mitchell, and W.~K. Durfee, ``Experimentally validated models
  for switching energy of low pressure drop digital valves for lightweight
  portable hydraulic robots,'' in \emph{{ASME}/{BATH} 2019 {Symposium} on
  {Fluid} {Power} and {Motion} {Control}, {FPMC} 2019, {October} 7, 2019 -
  {October} 9, 2019}, ser. {ASME}/{BATH} 2019 {Symposium} on {Fluid} {Power}
  and {Motion} {Control}, {FPMC} 2019.\hskip 1em plus 0.5em minus 0.4em\relax
  Longboat Key, FL, United states: American Society of Mechanical Engineers
  (ASME), 2020, p. Fluid Power Systems and Technology Division.

\bibitem{valworx_53685369_nodate}
\BIBentryALTinterwordspacing
Valworx, ``5368/5369 {Series} {Full} {Port} {Lead} {Free} {Brass} 3-{Way}
  {Ball} {Valves}.'' [Online]. Available:
  \url{https://s3.amazonaws.com/cdn.valworx.com/downloads/datasheets/valworx/53685369.pdf}
\BIBentrySTDinterwordspacing

\end{thebibliography}
